\pdfoutput=1
\documentclass[11pt]{article}
\usepackage[final]{neurips_2024}

\usepackage{times}
\usepackage{latexsym}
\usepackage[colorlinks=true,citecolor=blue]{hyperref}

\usepackage[T1]{fontenc}
\usepackage[utf8]{inputenc}
\usepackage{microtype}
\usepackage{inconsolata}

\usepackage{graphicx}

\usepackage{tcolorbox}
\tcbuselibrary{skins}
\usepackage{lipsum}
\usepackage{xcolor}
\usepackage{amssymb}
\usepackage{enumitem}
\usepackage{booktabs}
\usepackage{tabularx}
\usepackage{multirow}
\usepackage{amsmath}
\usepackage{tikz}
\usepackage{tikz-qtree}
\usetikzlibrary{calc,positioning}
\usepackage{relsize}
\makeatletter
\def\txds@@scale{1}
\input{ot1tx-ds.fd}
\makeatother
\usepackage[bb=tx]{mathalpha}
\usepackage[capitalize]{cleveref}
\usepackage{soul}
\usepackage{subcaption}
\usepackage{mathtools}
\usepackage{nicefrac}

\usepackage{enumitem} 
\usepackage{tabularx} 
\usepackage[parfill]{parskip}
\usepackage{tcolorbox}
\usepackage{makecell}

\newcommand{\inter}[1]{\ensuremath\lceil{#1}\rceil}
\newcommand{\ia}{\textit{inter alia}}
\newcommand{\num}[1]{\ensuremath{#1}}
\newcommand{\sstat}[2]{\hphantom{1}\num{#1}{\scriptsize{\textcolor{gray!70}{\num{\pm{}#2}}}}}
\newcommand{\bstat}[2]{\num{#1}{\scriptsize{\textcolor{gray!70}{$\pm${\num{#2}}}}}}

\newcolumntype{C}{>{\centering\arraybackslash}X}

\newcommand{\look}[1]{\underline{#1}}
\newcommand{\Look}[1]{\look{\textcolor{red}{#1}}}

\newcommand{\Path}{\ensuremath{\mathbb{P}}}
\newcommand{\zero}{\ensuremath{\mathbb{0}}}
\newcommand{\one}{\ensuremath{\mathbb{1}}}
\newcommand{\two}{\ensuremath{\mathbb{2}}}
\newcommand{\pcomp}{\mathbin +}
\newcommand\pinvt[1]{\mathop{-}#1}
\newcommand\pinv[1]{(\pinvt{#1})}

\newcommand{\perm}[1]{\ensuremath{\mathcal{S}_{#1}}}

\usepackage{bm}
\renewcommand{\arraystretch}{0.85}
\newcommand{\mtrx}[1]{{\bm{#1}}}
\newcommand{\eg}{\textit{e.g.},}
\newcommand{\ie}{\textit{i.e.},}
\newcommand{\ape}{\textsc{ape}}
\newcommand{\rope}{\textsc{r}\textnormal{o}\textsc{pe}}

\title{Algebraic Positional Encodings}

\usepackage{authblk}
\usepackage{varwidth}

\renewcommand*{\Affilfont}{\normalsize\normalfont}
\newcommand*{\Emailfont}{\small\normalfont\texttt}

\renewcommand*{\Affilfont}{\normalsize\normalfont}
\newcommand{\aalto}{1}
\newcommand{\uob}{2}
\newcommand{\yaiyai}{5}
\newcommand{\uog}{3}
\newcommand{\chalmers}{4}
\newcommand{\authmail}[3]{
\protect\begin{varwidth}[t]{\linewidth}\protect\centering
#1{\Affilfont\textsuperscript{#2}}
\par \Emailfont{#3}
\protect\end{varwidth}}

\author[ ]{\authmail{Konstantinos Kogkalidis}{\aalto,\uob}{kokos.kogkalidis@aalto.fi}}
\author[ ]{\authmail{Jean-Philippe Bernardy}{\uog,\chalmers}{jean-philippe.bernardy@gu.se}}
\author[ ]{\authmail{Vikas Garg}{\aalto,\yaiyai}{vgarg@csail.mit.edu}}
\affil[\aalto]{Aalto University}
\affil[\uob]{University of Bologna}
\affil[\uog]{University of Gothenburg}
\affil[\chalmers]{Chalmers University of Technology}
\affil[\yaiyai]{YaiYai Ltd}

\begin{document}
\maketitle
\begin{abstract}
We introduce a novel positional encoding strategy for Transformer-style models, addressing the shortcomings of existing, often \textit{ad hoc}, approaches. 
Our framework provides a flexible mapping from the algebraic specification of 
a domain to an interpretation as orthogonal operators. 
This design preserves the algebraic characteristics of the source domain, 
ensuring that the model upholds its desired structural properties. 
Our scheme can accommodate various structures, including sequences, grids and trees, as well as their compositions.
We conduct a series of experiments to demonstrate the practical applicability of our approach. 
Results suggest performance on par with or surpassing the current state-of-the-art, without hyper-parameter optimizations or ``task search'' of any kind.
Code is available through \url{https://aalto-quml.github.io/ape/}.
\end{abstract}

\section{Introduction}
Attention-based models inheriting from the Transformer~\citep{vaswani2017attention} have become ubiquitous in neural computation, supplanting the go-to models of the last decade and driving a continuous stream of breakthroughs across diverse domains. 
Their success is perhaps at odds with the Transformer's structural lenience -- its key building block, dot-product attention, is by default unable to perceive and utilize the structure and arrangement of the input/output tokens being processed.
To address this limitation, a plethora of works have sought to endow Transformers with appropriate inductive biases.
The most common strategy is to adjust token representations via so-called \textit{positional encodings}; vector operations that hint at the structure being modeled.
Nonetheless, most positional encoding schemes to date are either empirically motivated, or tailored to specific tasks.
This renders their theoretical evaluation challenging, and hinders any prospects of a unifying framework.

In this study, we seek to fill this gap with a theory-first approach.
Through the lens of group theory, we scrutinize some of the most commonly targeted data structures, and express them by means of inductive definitions that reveal and explicate their structural properties.
Leveraging this analysis, our modeling strategy invokes a homomorphic interpretation that maps each domain into \textbf{algebraic positional encodings} (\ape{}): attention-compatible vector operations parameterizing (subgroups of) the orthogonal group.
In the sequential context, algebraic positional encodings streamline the widely adopted rotary encodings of \citet{su2023roformer}, while also offering clear theoretical insights on their success.
More importantly, algebraic positional encodings naturally extend to non-sequential domains, such as $\kappa$-ary trees and multidimensional regular grids, paving the way for a simple and elegant methodology for interpretable and domain-general structurally-refined Transformers.
We carry out an experimental evaluation in settings that allow for reproducible and statistically sound conclusions.
Across the tasks considered, algebraic positional encodings consistently and significantly outperform strong baselines at an aggregate level, providing initial but compelling evidence that they constitute not just a \textit{sensible meta-theory} for positional encodings, but also an \textit{actionable alternative} to the current state of the art.

\begin{table}
    \centering\smaller
    \renewcommand{\arraystretch}{1.05}
    \begin{tabular}{l@{~}p{0.95\textwidth}@{}}
    \multicolumn{2}{c}{\textsc{Contributions \& Paper Structure}}\\
    \toprule
    \multicolumn{2}{@{}l}{\S\ref{sec:background} \textsc{Background}}\\
    \multicolumn{2}{@{}p{\textwidth}@{}}{We introduce the problem (\S\ref{sec:attention}) and the vocabulary of the solution (\S\ref{sec:group}).}\vspace{0.5em}\\
    \multicolumn{2}{@{}l}{\S\ref{sec:algebra} \textsc{Theory}}\\
    \multicolumn{2}{@{}p{\textwidth}@{}}{We provide an algebraic characterization of positions in the context of different ambient structures. We frame algebraic positional encodings as structure-preserving semantic interpretations, and present reference implementations. Concretely:}\\
        \S\ref{sec:sequences} & \textit{Sequences} are an isomorphism of the free group $\langle \one \rangle$ (\ie{} the integers, $\mathbb{Z}$), and can be interpreted as a single generator subgroup of the orthogonal group $O(d)$.\\
        \S\ref{sec:rope} & \textit{Rotary Positional Encodings} correspond to a (quite literally) special case of this interpretation: $SO(d)$.\\
        \S\ref{sec:trees} & \textit{k-ary Trees} are an isomorphism of the finitely generated group $\langle \one, \two \dots \kappa \rangle$, and can be interpreted as a finitely generated subgroup of $O(d)$.\\
        \S\ref{sec:grids} & \textit{Regular Grids} are the group direct sum of multiple sequences. They can be interpreted as the matrix direct sum of their components' interpretations.\\
        \S\ref{sec:variants} & \textit{Extensions} can be obtained in multiple directions.\vspace{0.5em}\\
    \multicolumn{2}{@{}l}{\S\ref{sec:experiments} \textsc{Practice}}\\
    \multicolumn{2}{@{}p{\textwidth}@{}}{We carry out fair and replicable experiments across all three structures analyzed, namely \textit{Sequence Transduction} (\S\ref{sec:strans}), \textit{Tree Transduction} (\S\ref{sec:treetrans}) and   \textit{Image Recognition} (\S\ref{sec:imagerec}), and find that algebraic positional encodings consistently match or outperform alternative schemes in the tasks considered (\S\ref{sec:results}).}\vspace{0.5em}\\
    \multicolumn{2}{@{}l}{\S\ref{sec:relwork} \textsc{Related Work}}\\
    \multicolumn{2}{@{}p{\textwidth}@{}}{We position our work in the context of the broader literature.}\\[0.5em]
    \multicolumn{2}{@{}l}{\S\ref{sec:limitations} \textsc{Limitations}}\\
    \multicolumn{2}{@{}p{\textwidth}@{}}{We close with a brief discussion of possible limitations and potential future improvements.}\\
    \bottomrule
    \end{tabular}
    \caption{A summary of this paper.}
    \label{table:summary}
\end{table}

\section{Background}\label{sec:background}
\subsection{The Problem with Dot-Product Attention}\label{sec:attention}
All transformer variants employ some variation of the multi-head scaled dot-product attention mechanism of \citet{vaswani2017attention}.
For each attention head, the dot-product attention between queries $\mtrx{X} \in \mathbb{R}^{m \times d}$ and keys $\mtrx{Y} \in \mathbb{R}^{n\times d}$ is defined as:
\begin{equation}
    \mathrm{atn}(\mtrx{X}, \mtrx{Y}) :=
        \mathrm{softmax}_{(n)}\!
        \left(
            \frac{
                (\mtrx{X}\mtrx{\Phi}^{(q)})
                (\mtrx{Y}\mtrx{\Phi}^{(k)})^\top
            }{
                \sqrt{d}
            }
        \right)
        \mtrx{Y}\mtrx{\Phi}^{(v)}
       \label{eq:atn}
\end{equation}
In equation (\ref{eq:atn}), matrices $\mtrx{\Phi}^{(q)}, \mtrx{\Phi}^{(k)}, \mtrx{\Phi}^{(v)} : \mathbb{R}^{d\times d}$ enact linear functions, applied point-wise (broadcasted) across all $m$ and $n$ entries of $\mtrx{X}$ and $\mtrx{Y}$. 
The dot-product term $(\mtrx{X}\mtrx{\Phi}^{(q)})(\mtrx{Y}\mtrx{\Phi}^{(k)})^\top$ contains unnormalized attention scores in the Cartesian product of queries and keys.
Unmodified, dot-product attention is permutation \textit{invariant} with respect to its second argument; 
that is, for any arbitrary permutation $p_n \in \perm{n}$:
\begin{equation}
 \mathrm{atn}(\mtrx{X}, \mtrx{Y}) \equiv \mathrm{atn}(\mtrx{X}, p_n(\mtrx{Y}))
\end{equation}
Unless one is dealing with orderless structures like multisets or fully connected graphs, this property is generally undesirable.
The lack of structural biases is typically counteracted by the component-wise addition of unidimensional periodic signals of varying frequencies.
These, however, often prove inadequate in data-scarce domains, where extensive pretraining is impossible, and structure-rich domains, where a sequence-of-tokens projection is too radical of a simplification.

\subsection{Recap on Group Theory}\label{sec:group}
To address this issue, we propose an algebraic treatment of positional encodings, based on principles lent from group theory.
For the sake of convenience and accessibility, we provide a brief recap of the notions of interest here.
A \textit{group} $G$ consists of a set of \textit{elements} and a \textit{binary operation} (\_$\cdot$\_) satisfying four fundamental laws:%
\begin{itemize}[topsep=0pt,leftmargin=*,noitemsep]
	\item The group is \textit{closed} under the the group operation. For all $a$, $b$ in $G$, $a\cdot b$ is also in $G$.
	\item The group operation is \textit{associative}. For all $a$, $b$, $c$ in $G$, $(a\cdot b)\cdot c = a\cdot (b\cdot c)$.
	\item The group operation has an \textit{identity} element $e$, such that for all $a$ in $G$, $a\cdot e = e\cdot a = a$.
	\item Each group member has an \textit{inverse}. For all $a$ in $G$, there exists some element $\overline{a}$ such that $a\overline{a} = \overline{a}a = e$, where $e$ is the identity element.
\end{itemize}
A group is characterized as \textit{finite} or \textit{infinite} depending on the number of elements it has.
If all elements of a group $G$ can be expressed as a combination of a subset $S$ of the group elements (combined by means of the group operation, applied either on the elements themselves or on their inverses), we write $G = \langle S \rangle$.
We say that $G$ is \textit{generated} by $S$, and we call the elements of $S$ the \textit{generators} of $G$.
A group with a single generator is called \textit{cyclic}.

\section{The Algebra(s) of Positions}
\label{sec:algebra}
Our objective is to establish a framework that offers general and extensible \textit{semantics} for positions across various structures -- what we commonly encounter in the literature as \textit{positional encodings}.
Most existing proposals adopt a rather parochial stance, relying on maneuvers or heuristics tailored to specific applications and driven, predominantly, by extensive empirical investigations. 
As such, they fall short with respect to accommodating or reflecting the properties of the underlying structure.
In this work, we follow a different approach.
We adopt Montague's perspective, succinctly paraphrased as:
\begin{quote}
``\textit{syntax is an algebra, semantics is an algebra, and meaning is a homomorphism between them}''~\citep{janssen2014foundations}.    
\end{quote}

We begin by noting that ``positions'' do not exist in isolation, but only in the context of some underlying ambient structure.
We contend that reasonable positional encodings (\textit{semantics}) may only be reliably obtained by taking into account exactly this structure, its formation rules and properties (\textit{syntax}), and then applying an appropriate interpretation (\textit{meaning}).
This is \textit{not} just an academic exercise: a careful syntactic specification is a prerequisite if we aim for semantics that adhere to certain properties, which is arguably preferable to searching for these properties in the wild.

\subsection{Sequences}
\label{sec:sequences}
\paragraph{Syntax}
We start from the simplest structure, and incidentally also the most standard one: the sequence.
The full range of positions a token can occupy within a sequence coincides exactly with the naturals, $\mathbb{N}$.
Relative paths \Path{} between any two positions can then be seen as the integers, $\mathbb{Z}$,
with positive (resp. negative) numbers denoting forward (resp. backward) offsets.
Using this insight, it is handy to inspect how the standard inductive definition of the integers provides the building blocks for path formation.
We start with two constants: the empty path ($\zero$), which relates any given point to itself, and the unit path ($\one$), which relates any point to its immediate next. 
We may compose simple paths into complex ones with the aid of a binary operation $\pcomp_{\Path}$.
This already suffices to specify all forward offsets.
In order to construct backward offsets, we need a unary operation $\pinv{}_{\Path}$, such that $\pinvt \rho$ denotes the inverse of $\rho$.
We can summarize the above by the grammar:
\begin{align}
    \Path := \zero ~ | ~ \one ~ | ~ \Path \pcomp_{\Path} \Path ~ | ~ \pinvt{\Path}
    \label{def:spath}
\end{align}
For this to make sense, the operations must be \textit{coherent}; that is, all ways to start from point $\rho_1$ and end up in point $\rho_2$ should be equivalent, even if apparently distinct.
The needed equivalences exactly correspond to the group laws, with closure internalized by the inductive definition of (\ref{def:spath}):
\begin{align}
	(\rho_1 \pcomp_{\Path} \rho_2) \pcomp_{\Path} \rho_3 &= \rho_1 \pcomp_{\Path} (\rho_2 \pcomp_{\Path} \rho_3)
    \tag{L1}
    \label{prop:assoc_of_plus}\\
    \rho \pcomp_{\Path} \zero &= \rho = \zero \pcomp \rho
    \tag{L2}
    \label{prop:id_of_plus}\\
    \rho \pcomp_{\Path} \pinv \rho &= \zero
    \tag{L3}
    \label{prop:def_of_inv}
\end{align}
The (unsurprising) insight here is that paths in a sequence form a free group, generated by a single generator ($\one$) -- the uniqueness of the generator exceptionally also makes the group abelian (\ie{} commutative). 
For convenience, we adopt the notational shorthand $\one^p$, where:
\begin{equation}
  	\one^{p} :=         \begin{cases}
    	\underbrace{\one \pcomp_{\Path} \cdots \pcomp_{\Path} \one}_{p} & p\geq 0\\
        \underbrace{\pinv \one \pcomp_{\Path} \cdots \pcomp_{\Path} \pinv \one}_{-p}) & p<0 
	\end{cases}
\end{equation}

\paragraph{Semantics}
\label{sec:seq_semantics}
The syntactic specifications of the previous paragraph impose constraints on the candidate semantic targets.
Among these candidates, we isolate and focus on $\langle \mtrx{W} \rangle$, the subgroup of the orthogonal group $O(d)$ that is generated by a single orthogonal matrix $\mtrx{W}$.
This semantics is not only sound%
\footnote{It is also complete except for the odd case where $\mtrx{W}^p=\mtrx{I}$ for some $p$. In practice, this kind of periodic behaviour does not arise randomly, and we can think of $\langle \mtrx{W} \rangle$ as being \textit{isomorphic} to $\Path$.}
with respect to the structure under scrutiny, but also a familiar object in machine learning literature~\cite[\ia]{arjovsky2016unitary,bernardy2022unitary}.
Note that for $\langle \mtrx{W} \rangle$, the group axioms are obtained for free from the orthogonal group, and the additional requirement of commutativity is again satisfied by the uniqueness of the generator.

To illustrate the correspondence between the two structures (and at risk of being pedantic), we spell out the homomorphism $\inter{.}$,
which maps paths $\Path$ to elements of $\langle \mtrx{W} \rangle$, and path operations to operations on orthogonal matrices of size $d$.
For the primitives, we have $\inter{\zero} := \mtrx{I}_{d}$ and $\inter{\one} := \mtrx{W}$.
Path composition amounts to matrix multiplication, \ie{} $\inter{\rho_1 \pcomp_{\Path} \rho_2} := \inter{\rho_1}\inter{\rho_2}$, while path inversion corresponds to matrix transposition, \ie{} $\inter{\pinvt \rho} := \inter{\rho}^{-1} \equiv \inter{\rho}^{\top}$.
The fact that orthogonal matrices form a group under multiplication is folklore; one can easily verify that the group laws hold also for the semantics.%
\footnote{The story is no different for $\mtrx{W}$ unitary, with the group structure provided by the unitary group $U(d)$, and path inversion interpreted as the matrix conjugate transpose.
}

\paragraph{Implementation}\label{sec:seq-implementation}
In practice, we have $\inter{\one^{p}} \mapsto \mtrx{W}^p$; a norm-preserving bilinear form $\mathbb{R}^d \times \mathbb{R}^{d} \to \mathbb{R}$ which can be used to mediate the dot-product between a query $q$ and a key $k$ offset by a relative distance of $p$.
The representation of all paths up to length $p$ can thus be implemented as a matrix collection $[\mtrx{W}^{0},\dots ,\mtrx{W}^{p}]$, which can asymptotically be obtained using $\mathcal{O}(\lceil \mathrm{log}_2(p) \rceil)$ matrix products (of exponentially larger matrices), and taking up the storage space equivalent of $(pd^2)$ floats.
Transposed, the same matrices also serve to represent backwards paths $[\mtrx{W}^{-p},\dots,\mtrx{W}^{0}]$.
Storing the representations of all relative paths between queries and keys in a tensor $\mtrx{T} : \mathbb{R}^{m\times n\times d\times d}$, we may then substitute the dot-product term of equation (\ref{eq:atn}) for the tensor contraction:
\begin{equation}
    \sum_{\alpha,\beta} \mtrx{X}^{}_{m\alpha} \mtrx{\Phi}^{(q)}_{\alpha\beta} \mtrx{T}^{}_{mn\beta\gamma} \mtrx{Y}^{}_{n\delta}\mtrx{\Phi}^{(k)}_{\delta\gamma}
    \label{eq:bad}
\end{equation}
Albeit transparent, this reduction strategy is computationally unappealing due to the doubly quadratic nature of $\mtrx{T}$.
We can do better by noting that $\mtrx{T}_{mn}$ is (definitionally) equal to:
\begin{equation}   
    \mtrx{T}_{mn\alpha\beta} = \sum_{\gamma} \mtrx{A}^{(X)}_{m\gamma\alpha} \mtrx{A}^{(Y)}_{n\gamma\beta}
\end{equation}
where $\mtrx{A}^{(X)}$ and $\mtrx{A}^{(Y)}$ are the matrices containing representations for the \textit{absolute} positions of the entries in $\mtrx{X}$ and $\mtrx{Y}$, respectively.
Concretely, a single relative representation is built by composing the \textit{inverted} representation of the source with the representation of the target.
Intuitively, each query follows the path that takes it \textit{back} to the origin, which then allows it to directly combine with each forward-offset key; see Figure~\ref{fig:sequential} for a visual example.
This insight allows us to keep the memory footprint of equation (\ref{eq:atn}) unchanged, replacing expression (\ref{eq:bad}) with:
\begin{equation}
    \sum_{\alpha,\beta,\gamma,\delta,\epsilon}
        \mtrx{X}^{}_{m\alpha} \mtrx{\Phi}^{(q)}_{\alpha\beta} 
        \mtrx{A}^{(X)}_{m\gamma\beta}
        \mtrx{A}^{(Y)}_{n\gamma\delta}
        \mtrx{Y}^{}_{n\epsilon}
        \mtrx{\Phi}^{(k)}_{\epsilon\delta}
    \label{eq:good}
\end{equation}
This version decomposes the tensor contraction into two matrix multiplications, essentially transforming (rotating or reflecting) the entries of $\mtrx{X}$ and $\mtrx{Y}$ independently according to their positions.

\subsection{Intermezzo: Equivalence with \rope{}}
\label{sec:rope}
The story so far should be reminiscent of the rotary positional encoding scheme of \citet[\rope{}]{su2023roformer}.
Not unlike our approach, \rope{} substitutes the vanilla dot-product for a position-dependent bilinear form.
Underlying the form is a $d\times d$-dimensional matrix $\mtrx{R}$ with a block-diagonal structure, where each $2\times 2$-sized block corresponds to a rotation matrix that acts on a $2$-dimensional subspace of $\mathbb{R}^d$.
These independent rotations are parameterized by a (fixed) set of base angles $\Theta := [\theta_1, \dots,\theta_{d/2}]$.
To incorporate position-dependence, \ie{} for a query/key pair at a relative distance of $p$, the base angles are multiplied by $p$, effectively altering the rotations applied.

At first glance, rotary encodings appear to be under-parameterized, and thus strictly weaker than orthogonal ones.
However, any orthogonal matrix $\mtrx{W} \in O(d)$ admits a canonical form $\mtrx{W} = \mtrx{P}\mtrx{Q}\mtrx{P}^\top$, where $\mtrx{P}$ is an orthogonal change of basis, and $\mtrx{Q}$ is block-diagonal, with the $2\times 2$-sized blocks being, once again, $2-$dimensional rotation matrices~\citep{murnaghan1931canonical}%
\footnote{We alert the reader that a \textit{constructive} proof of this decomposition has proven surprisingly difficult to find.}.
Owing to the orthogonality of $\mtrx{P}$, raising $\mtrx{W}$ to its $p$th power is equal to $\mtrx{P}\mtrx{Q}^p\mtrx{P}^\top$ (\ie{} it leaves the change of basis unaffected).
In turn, raising $\mtrx{Q}$ to its $p$th power is equivalent to simply multiplying the rotation angles of its blocks by $p$.
Finally, given the linearity of the transformations $\mtrx{\Phi}^{(q)}$ and $\mtrx{\Phi}^{(k)}$, their compositions with $\mtrx{P}$ are also linear.
By identifying $\mtrx{Q}$ with \rope{}'s $\mtrx{R}$, we can then see that, for any given collection of angles $\Theta$, \ape{} and \rope{} coincide under the substitutions:
\begin{equation}
    \begin{array}{ccc}
         \mtrx{\Phi}^{(q)}_{\textnormal{\rope}} = \mtrx{\Phi}^{(q)}_{\vphantom{\rope}}  \mtrx{Q} & 
         \text{and} &
         \mtrx{\Phi}^{(k)}_{\textnormal{\rope}} = \mtrx{\Phi}^{(k)}_{\vphantom{\rope}}  \mtrx{Q}
    \end{array}
\end{equation}
In other words, \emph{\ape{} is practically equivalent to a \textit{trainable} version of \rope{}, where the rotation angles $\mathit{\Theta}$ may vary and be optimized during training}%
\footnote{An alternative reading is that even though orthogonal matrices are generally more expressive than rotation matrices (allowing not just rotations but also reflections), the Transformer's architecture makes up for \rope{}'s reduced expressivity by supplying a free change of basis through its trainable weights $\mtrx{\Phi}$.}.

Which of the two parameterizations is preferable is up to debate.
On the one hand, \ape{}'s formulation is FLOP-optimized (being just matrix multiplications), and obviates the need for backpropagating through trigonometric functions (which are periodic, non-monotonic, and prone to gradient instabilities).
On the other hand, \rope{}'s diagonalized form gives access to a memory-efficient contraction that does away with the matrix multiplications of expression (\ref{eq:good}) altogether; we direct the interested reader to \citet[Section 3.4.2]{su2023roformer} for a reference implementation%
\footnote{For more practical insights on initializing and parameterizing \ape{} and translating between \ape{} and \rope{}, please refer to Appendix~\ref{appendix:init}.}.

In either case, the equivalence between the two is confined to the \textit{sequential} setup; we will now move on to generalize our strategy to other, \textit{previously inaccessible}, structures.

\subsection{Trees}\label{sec:trees}
\paragraph{Syntax}
In the previous section, we characterized the structure of relative paths on a sequence as the free group with one generator, and uncovered a (practically) isomorphic interpretation in the subgroup of orthogonal matrices with a single generator.
Upon closer inspection, we note that a sequence can be viewed as a special case of the more general structure of $\kappa$-ary branching trees, where the branching factor $\kappa$ just so happens to be 1.
Denoting the more general case as $\Path_{\kappa}$, we must first extend the set of primitives to include all branching options, $\one, \two, \dots \mathbb{\kappa} : \Path_{\kappa}$.
Each primitive now denotes a choice of branch (except for $\zero$, which is again the empty path).
Paths now form a free group with $\kappa$ distinct generators.
The presence of multiple generators means that commutativity no longer holds; $\one \pcomp_{\Path_{\kappa}} \two$ is distinct from $\two \pcomp_{\Path_\kappa} \one$ (the former prescribes a descent down branch $\one$ then branch $\two$, whereas the latter prescribes a descent down branch $\two$ then branch $\one$).
Inversion is as before: for every path from each local root to some descendant down the line, there is also an inverse path from that descendant up to its ancestor.
Perhaps more interestingly, upwards and downwards paths can be joined, allowing the precise specification of relative paths between any two nodes, even when the two do not share a single line of descent (think nephews, aunts and all other sorts of distant relatives, see Figure~\ref{fig:tree} for an example).
Adjusting grammar~(\ref{def:spath}) accordingly, we have:
{\lineskip=0pt
\begin{align}
    \Path_{\kappa} := \zero ~ | ~ \one ~ | ~ \two ~ | \dots ~ | ~ \kappa ~ | ~ \Path_{\kappa} \pcomp_{\Path_{\kappa}} \Path_{\kappa} ~ | ~ \pinvt {\Path_{\kappa}}
    \label{def:tpath}
\end{align}
}
with laws \ref{prop:assoc_of_plus}, \ref{prop:id_of_plus} and \ref{prop:def_of_inv} still in effect.

\paragraph{Semantics}
The interpretation follows along the same lines as before.
This time around, however, we cannot make do with a single orthogonal matrix $\mtrx{W}$ -- we need a collection of $\kappa$ matrices, one for each branch option.
As a consequence, the semantic target is now $\langle \mtrx{W}_1, \mtrx{W}_2, \dots \mtrx{W}_\kappa \rangle$.
Note that the target is no longer commutative (in alignment with the source).

\paragraph{Implementation}
For a tree structure of depth $\delta$ and branching factor $\kappa$, let $\nu$ denote the number of \textit{unique} absolute positions occupied (upper bound by $\kappa^\delta$ in the case of a complete tree).
Their representations can be computed in $\delta \kappa$ steps of parallel matrix-matrix multiplications and a memory cost of $\nu{}d^2$, as follows.
First, we can build up a collection of all unique absolute paths, each represented as a (right-padded) word of length $\delta$ from the vocabulary of primitives.
Their corresponding representations constitute a tensor of size $\nu{}\times d \times d$, initialized as $\nu$ identity matrices.
We can then iterate across these words in parallel, one primitive per step (\ie{} depth) $t$, 
selecting all words that take the same branching direction at the current depth, and right-multiplying their representations by the corresponding orthogonal generator.
Finally, absolute paths can be composed into relative ones using the modified dot-product attention of expression (\ref{eq:good}), just like before.

\subsection{Grids}%
\label{sec:grids}
The generalization from sequences to trees rests on the observation that a sequence is a tree with a deficit of choices.
An altogether different axis of generalization can be obtained by recalling that composite groups can be constructed by joining together two or more elementary groups.
Moreover, if it just so happens that the original groups were abelian, then so is their composition; in that case, we call the composite a \textit{group direct sum}.
This construction provides access to an extension from sequences to multidimensional regular grids.

For the sake of simplicity and without loss of generality, we consider a standard instance of a two-dimensional grid: an image.
An image is a collection of pixels (or pixel patches) that inhabit a coordinate system $(h, w)$.
Each of $h$ and $w$ is the product of grammar~(\ref{def:spath}), inheriting all path-related notions discussed earlier.
Since $\Path$ is an abelian group, the coordinate system also constitutes an abelian group $\Path^2 := \Path \oplus \Path$.
The new group and inversion operations are $\pcomp_{\Path^{2}}$ and $\pinv{}_{\Path^2}$, and denote the act of joining and inverting two-dimensional paths, respectively.
Both are canonically defined component-wise, on the basis of their one-dimensional counterparts:
\begin{align}
    (x, y) \pcomp_{\Path^2} (z,w)    &:= (x \pcomp_{\Path} y, z\pcomp_{\Path} w)\\
    \pinvt {(x,y)}                  &:= (\pinvt x, \pinvt y)
\end{align}
with $\zero^2 := (\zero, \zero)$ as the new neutral element.
Intuitively, $\pcomp_{\Path^2}$ corresponds to vector addition, and $\pinv{}_{\Path^2}$ to a reflection about the origin with respect to both axes.

\paragraph{Semantics}
The specifications above allow us to reuse the notions from Section~\ref{sec:seq_semantics} in order to interpret the components and operations of $\Path^2$.
What is left unspecified is the interpretation of the group elements themselves; that is, we have yet to explicate what an object of $\inter{\Path \oplus \Path}$ looks like.
The quest is a short one; the notion of a direct sum carries over to matrices, where it is defined as:
\begin{align}
 \mtrx{A} \oplus \mtrx{B} &:= 
    \begin{bmatrix}
        \mtrx{A} & \mtrx{0}\\
        \mtrx{0} & \mtrx{B} 
    \end{bmatrix}
\end{align}
From this, we get the (rather straightforward) interpretation $\inter{(\rho_1, \rho_2)} \mapsto \inter{\rho_1} \oplus \inter{\rho_2}$.

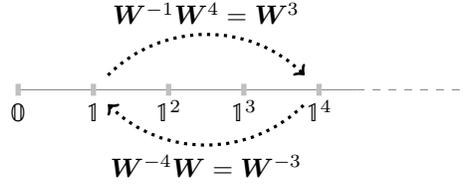
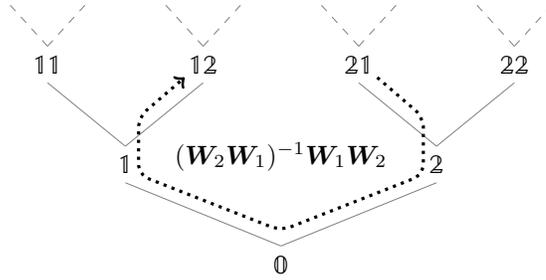
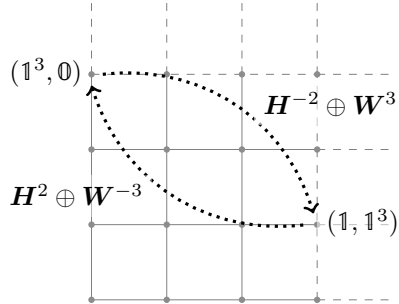
\begin{figure}
    \tikzset{
        grid/.style={very thin,gray}
    }
    \tikzset{
        dot/.style={
            draw,
            gray!50,
            line width=1pt,
            minimum height=3.5pt,
            inner sep=0pt,
            anchor=center,
            outer sep=5pt}
    }
    \tikzset{
        label/.style={text height =1.5ex}
    }
    \begin{subfigure}{\textwidth}
    \centering
    \begin{tikzpicture}
        \draw[grid] (0,0) -- (4.5,0);
        \draw[grid,dashed] (4.5,0) -- (6,0);
        \node[dot] (0) at (0,0) {};
        \node[below=-5pt of 0,label] {$\zero$};
        \node[dot] (1) at (1,0) {};
        \node[below=-5pt of 1,label] {$\one$};
        \node[dot] (2) at (2,0) {};
        \node[below=-5pt of 2,label] {$\one^{2}$};
        \node[dot] (3) at (3,0) {};
        \node[below=-5pt of 3,label] {$\one^{3}$};
        \node[dot] (4) at (4,0) {}; 
        \node[below=-5pt of 4,label] {$\one^{4}$};
        
        \draw[->,dotted,very thick] (1) to[bend left=45] node[midway, above]{$\mtrx{W}^{-1}\mtrx{W}^4 = \mtrx{W}^{3}$} (4);
        \draw[<-,dotted,very thick] (1) to[bend right=45] node[midway, below]{$\mtrx{W}^{-4}\mtrx{W} = \mtrx{W}^{-3}$} (4);
    \end{tikzpicture}
    \caption{The half-axis of absolute positions on a sequence, with a visualization of the two directions of relative paths between points 1 and 4. In either case, the interpretation is the matrix multiplication of the inverted source against the target.}
    \label{fig:sequential}
    \end{subfigure}\vspace{0.25\baselineskip}
    \begin{subfigure}{\textwidth}
    \centering
    \hspace{-2.5em}\begin{tikzpicture}[grow'=up,anchor=center,sibling distance=0.8cm,level distance=1.33cm]
    \tikzset{level 3/.style={sibling distance=13pt}}
    \tikzset{level 3/.style={level distance=30pt}}
    \tikzset{edge from parent/.style={grid,draw}}
    \tikzset{every tree node/.style={label}}
            \Tree 
            [.\node (1) {$\zero$};
                [.\node (2) {$\one$};
                    [.\node (4) {$\one\one$};
                        \edge[dashed]; {}
                        \edge[dashed]; {}
                    ]
                    [.\node (5) {$\one\two$};
                        \edge[dashed]; {}
                        \edge[dashed]; {}
                    ]
                ]
                [.\node (3) {$\two$};
                    [.\node(6) {$\two\one$};
                        \edge[dashed]; {}
                        \edge[dashed]; {}
                    ]
                    [.\node(7) {$\two\two$};
                        \edge[dashed]; {}
                        \edge[dashed]; {}
                    ]
                ]
            ]  
        \draw [->,dotted,very thick,rounded corners,transform canvas={shift={(0,5pt)}}]
            ($(6.south) + (7pt, -3pt)$) --
            ($(3.north) + (-5pt,6pt)$) -- 
            ($(3.south) + (-5pt,-2pt)$) -- 
            ($(1.north) + (0, 0.8pt)$) -- 
            ($(2.south) + (5pt,-2pt)$) -- 
            ($(2.north) + (5pt,6pt)$) -- 
            ($(5.south) + (-7pt,-3pt)$);
        \node (x) at (0pt, 41pt) {$(\mtrx{W}_2\mtrx{W}_1)^{-1}\mtrx{W}_{1}\mtrx{W}_2$};
    \end{tikzpicture}
    \caption{The space of paths on binary branching trees, with an illustration of the relative path from $\two\one$ to $\one\two$. Same as before, the interpretation is the matrix multiplication of the inverted source against the target}
    \label{fig:tree}
    \end{subfigure}\vspace{0.25\baselineskip}
    \begin{subfigure}{1\textwidth}
    \centering
    \hspace{-3em}\begin{tikzpicture}

        \draw[step=1cm,grid] (0,0) grid (2.99,2.99);
        \draw[dashed,grid] (0, 3) -- (3, 3);
        \draw[dashed,grid] (3, 0) -- (3, 3);
        \draw[dashed,grid] (0, 3) -- (0, 4);
        \draw[dashed,grid] (1, 3) -- (1, 4);
        \draw[dashed,grid] (2, 3) -- (2, 4);
        \draw[dashed,grid] (3, 3) -- (3, 4);
        \draw[dashed,grid] (3, 0) -- (4,0);f
        \draw[dashed,grid] (3, 1) -- (4,1);
        \draw[dashed,grid] (3, 2) -- (4,2);
        \draw[dashed,grid] (3, 3) -- (4,3);

        \foreach \x in {0,1,2,3} {
        \foreach \y in {0,1,2,3} {
            \filldraw[gray!90] (\x,\y) circle (1pt);
           }
        }
        \node[left, label,fill opacity=0.5, text opacity=1.] at (0, 3) {$(\one^3, \zero)$};
        \node[right, label,fill opacity=0.5, fill=white, text opacity=1.] at (3,1) {$(\one, \one^3)$};
        \draw[->, dotted, very thick]
            (0.15, 3) 
            to[bend left=40] 
            node[midway,above,right,fill opacity=0.6, fill=white, text opacity=1.,xshift={0.25cm}]{$\mtrx{H}^{-2}\oplus\mtrx{W}^3$} 
            (2.95, 1.15);
        \draw[<-, dotted, very thick] 
            (0, 2.85) 
            to[bend right=40] 
            node[midway,below,left,fill opacity=0.4, fill=white, text opacity=1., xshift={-0.25cm}]{$\mtrx{H}^{2}\oplus\mtrx{W}^{-3}$} 
            (2.85,1);
                
    \end{tikzpicture}
    \caption{The quarter-plane of absolute positions on a 2-dimensional grid, with a visualization of the two directions of relative paths between points $(3,0)$ and $(1,3)$. The interpretation is now a block-diagonal matrix consisting of the blocks interpreting the path over each coordinate.}
    \label{fig:grid}
    \end{subfigure}
    \caption{Example paths and their interpretations across the structures examined.}
    \label{fig:images}
\end{figure}

\paragraph{Implementation} 
In practice, we now split the vector space in two independent parts.
The first part is modulated by orthogonal matrices from $\langle \mtrx{H} \rangle$, and the second part by orthogonal matrices from $\langle \mtrx{W} \rangle$.
For a query $q$ and a key $k$ that reside at a relative distance of $(h, w)$, their attention score is computed as $q(\mtrx{H}^h \oplus \mtrx{W}^w)k$ -- see Figure~\ref{fig:grid} for an illustration.
Each axis contributes an additive but separable factor to the attention score, forcing the model to learn contextual alignments between token pairs on the basis of their coordinate-wise distances.
Not much else is different: we can still compute all matrices in parallel, temporally bound by a logarithmic complexity of $\mathrm{log}_2(\mathrm{max}(h, w))$ and $\mathrm{max}(h,w)(\frac{d}{2})^2$ storage space, given a grid of size $(h,w)$.
Subquadratic memory complexity can once more be achieved by virtue of diagonalization, just as in the sequential case.

\subsection{Variants \& Extensions}\label{sec:variants}
The structures that we have seen so far are not the only ones that our methodology can tackle -- in fact, many other group-like structures are amenable to similar interpretations.
We sketch out some enticing examples below.

\paragraph{Absolute Positions}
\label{sec:absolute}
Our analysis has so far focused on paths \textit{relative} to positions.
Fixing the point of origin allows a straightforward simplification to \textit{absolute} positions.
The new structure is that of a \textit{monoid}: there's no longer an inversion, and laws \ref{prop:assoc_of_plus} and \ref{prop:id_of_plus} only are now in effect.
The framework remains largely unchanged: one can still use subgroups of matrices to represent positions, except this time applying them on either the queries or the keys (rather than both).

\paragraph{Periodic Domains}
\label{sec:periodic}
Under addition, the integers form an \textit{infinite} cyclic group.
An interesting twist would be to consider the positional encodings of \textit{finite} cyclic groups instead.
Such structures are not uncommon; in chemistry, for instance, a benzene molecule comprises six carbon atoms arranged in a ring.
The semantics of such a structure would need to be of a matching period; that is, we would need a generator $\mtrx{W}$ such that $\mtrx{W}^6 = \mtrx{I}$.
Such a parameterization is straightforward; we simply need to fix the orthogonal matrix so as to have it implement rotations at angle-multiples of $\pi / 3$.

\paragraph{Time Series \& Subsampling}
Our sequential case analysis assumed a dense sequence with a uniform sampling rate. However, our strategy also applies to any series, even if sparsely sampled, as long as the sampling rate is quantized (\ie{} a multiple of some constant step). That is, positional indices (and their representations) do not need to match the placement of tokens in the sequence.

\paragraph{Composite Groups}
The direct sum interpretation of Section~\ref{sec:grids} is applicable for arbitrary groups that can be described as products, commutative or otherwise.
This allows the representation of positional encodings for several other kinds of composite structures that can be concocted using the same principles, such as sequences of trees, trees of grids, etc.

\paragraph{Beyond Dot-Product Attention}
Throughout the previous sections, we have adopted a dot-product formulation for the attention weight function.
Nonetheless, \ape{} can be readily integrated into any other attention mechanism, such as linear~\citep{katharopoulos2020transformers}, cluster~\citep{vyas2020fast} and ``softmax-free''~\citep{lu2021soft} variants, \textit{inter alia}.

\section{Experiments}\label{sec:experiments}
To assess the viability of our approach, we conduct a series of experiments across a range of tasks, in setups that allow for replicable and reliable comparisons with alternatives.
When using \ape{}, we follow \citet{wu2021transformer} in scaling the dot-product score between two tokens at a distance of $p$ (\ie{} $p$ steps away) by $p^c$; here, we set $c := 0.98$.
This serves to stabilize training by introducing a locality bias (or long-distance decay) factor.
For the sake of parameter compression, we share the orthogonal matrices between the different encoder/decoder layers, but use a distinct matrix (or collection of matrices) per head.
To isolate and quantify the effect of initialization, we report results on two different initialization strategies: one where the orthogonal operators are set to mimic \rope{} rotations (default), and one where they are set to be close to the identity (no init).
Similarly, to isolate and quantify the effect of trainability when comparing to \rope{}, we report results over both fixed (frozen) and trainable (tuned) rotation angles.

We provide an extensive account of our experimental setups in Appendix~\ref{appendix:esetup}.

\subsection{Sequence Transduction}\label{sec:strans}
\paragraph{Machine Translation}
First, we follow \citet{vaswani2017attention} in training a Transformer\textsubscript{\textsc{base}} model on machine translation over \textsc{wmt14 en$\to$de}~\citep{bojar2014findings}.

To provide a comprehensive comparison, we pit our proposed methodology against standard positional encoding schemes from the literature: the vanilla \textit{Sinusoidal} encodings of ~\citet{vaswani2017attention}, the \textit{Absolute} encodings of \citet{gehring2017convolutional}, the \textit{Relative} encodings of \citet{shaw2018self} and the \textit{Rotary} encodings of \citet{su2023roformer}.
To ensure a fair comparison, we allow all models the exact same budgets (both memory and time).

\paragraph{Synthetic Tasks}
We further examine three standard sequence transduction tasks: sequence copying, sequence reversal, and sequence repetition. 
These are meant to directly assess each model's capacity for algorithmic induction, in setups where explicit position-based addressing, both absolute and relative, is required.

\newcommand{\model}[1]{\textit{#1}}

\begin{table}
    \centering
    \begin{subtable}{1\textwidth}
    \centering
    \smaller
    \begin{tabularx}{\textwidth}{
        @{~}l@{\quad}
        C
        C
        C
        C
        C
        C
        C
        @{~~}
    }
 		& \multicolumn{7}{c@{}}{Scheme}\\
		\cmidrule{2-8}
		\multicolumn{1}{@{}l}{\multirow{2}{*}{Task}}
		& \model{Sinusoidal}
		& \model{Absolute}
		& \model{Relative}
        & \multicolumn{2}{c}{\model{Rotary}}
        & \multicolumn{2}{c}{\model{Algebraic}}
    \\
    & & & & {\model{(frozen)}} & {\model{(tuned)}} & \model{(/w init)} & \model{(w/o init)}
    \\ 
		\toprule
		\textsc{wmt14 en$\to$de} 
            & \bstat{14.57}{0.12} 
            & \bstat{22.09}{0.11} 
            & \bstat{23.15}{0.03} 
            & \bstat{\Look{24.03}}{0.06}
            & \bstat{\look{23.92}}{0.20}
            & \bstat{\look{23.93}}{0.10}
            & \bstat{{23.84}}{0.10}
        \\
		\multicolumn{1}{r}{\smaller\textcolor{gray}{(\textsc{bleu} / $\uparrow$)}}\\[-2pt]
		\addlinespace
		\textsc{Copy}
			& \sstat{1.01}{0.00}
			& \sstat{1.11}{0.00}
			& \sstat{\Look{1.00}}{0.00}
			& \sstat{\Look{1.00}}{0.00}
            & \sstat{\Look{1.00}}{0.00}
			& \sstat{\Look{1.00}}{0.00}
            & \sstat{\Look{1.00}}{0.00}
		\\
		\textsc{Repeat}
			& \sstat{1.85}{0.15}
			& \sstat{3.66}{0.06}
			& \sstat{1.44}{0.16}
			& \sstat{\look{1.08}}{0.12}
            & \sstat{\Look{1.00}}{0.00}
            & \sstat{\Look{1.00}}{0.00}
            & \sstat{1.02}{0.00}
		\\
		\textsc{Reverse}
			& \sstat{3.92}{0.99}
			& \sstat{4.62}{0.67}
			& \sstat{4.08}{1.12}
			& \sstat{1.09}{0.02}
            & \sstat{\Look{1.01}}{0.00}
            & \sstat{\Look{1.01}}{0.00}
            & \sstat{\look{1.03}}{0.02}
		\\[-3pt]
		\multicolumn{1}{r}{\smaller\textcolor{gray}{(\textsc{ppl.} / $\downarrow$)}}\\
		\bottomrule
    \end{tabularx}
    \caption{Performance results on neural machine translation and synthetic sequence transduction.}
    \label{tab:seq_results}
    \end{subtable}\vspace{0.25\baselineskip}
    \begin{subtable}{1.005\textwidth}
    \centering
    \smaller
    \begin{tabularx}{\textwidth}{@{~}l@{~}
                    c@{\hskip 0.5pt}c@{\hskip 5pt}
                    c@{\hskip 0.5pt}c@{\hskip 5pt}
                    c@{\hskip 0.5pt}c@{\hskip 5pt}
                    c@{\hskip 2.5pt}c@{~}}
        & \multicolumn{8}{@{}c@{}}{Task/Regression}\\
        \cmidrule{2-9}
        & \multicolumn{2}{c@{\qquad}}{\textsc{Copy}} 
        & \multicolumn{2}{c@{\qquad}}{\textsc{Rotate}} 
        & \multicolumn{2}{c@{\qquad}}{C\textsubscript{3}}
        & \multicolumn{2}{c}{\textsc{Tree-Ops}}\\
        \cmidrule(l{3pt}r{3pt}){2-3}
        \cmidrule(l{3pt}r{3pt}){4-5}
        \cmidrule(l{3pt}r{3pt}){6-7}
        \cmidrule(l{3pt}r{3pt}){8-9}
        \multicolumn{1}{@{}l}{Scheme} & breadth & depth & breadth & depth & breadth & depth & breadth & depth \\
        \toprule
        \textit{Sinusoidal}
        		& \sstat{1.06}{0.01} & \sstat{5.68}{0.63}
        		& \sstat{6.93}{0.38} & \sstat{7.13}{0.35}
        		& \sstat{2.66}{0.10} & \sstat{2.78}{0.08}
        		& \bstat{20.53}{7.11} & \bstat{64.86}{6.41}
        \\
        \textit{Tree-SQ}
                & \sstat{1.29}{0.01} & \sstat{1.07}{0.00}
                & \sstat{2.60}{0.16} & \sstat{1.87}{0.24}
                & \sstat{2.27}{0.59} & \sstat{2.29}{0.24}
                & \bstat{19.18}{3.23} & \bstat{16.41}{6.14}
        \\
        \textit{Absolute}
        		& \sstat{6.64}{0.12} & \sstat{7.02}{0.17}
                & \sstat{7.77}{0.15} & \sstat{7.24}{0.20}
        		& \sstat{2.77}{0.21} & \sstat{2.79}{0.22}
                & \bstat{37.78}{0.72} & \bstat{48.91}{5.83}
        \\
        \textit{Relative}
                & \sstat{1.01}{0.00} & \sstat{6.12}{0.06}
                & \sstat{6.00}{0.25} & \sstat{7.72}{0.28}
                & \sstat{1.70}{0.07} & \sstat{2.43}{0.04}
                & \sstat{2.36}{0.02} & \bstat{16.86}{1.27}
        	\\
        \textit{Rotary (frozen)}
                & \sstat{\look{1.42}}{0.58} & \sstat{2.46}{0.59}
                & \sstat{4.58}{0.30} & \sstat{4.97}{1.79}
                & \sstat{1.55}{0.34} & \sstat{2.15}{0.22}
                & \sstat{2.53}{0.08} & \bstat{33.54}{9.04}
        \\
        \textit{Rotary (tuned)}
        		& \sstat{\Look{1.00}}{0.00} & \sstat{1.70}{0.05}
        		& \sstat{4.07}{0.34} & \sstat{2.60}{0.11}
        		& \sstat{1.08}{0.02} & \sstat{1.90}   {0.22}
        		& \sstat{2.55}{0.05} & \bstat{20.87}{0.33}
        \\
        \addlinespace
        \textit{Algebraic (\textbf{seq})}
                & \sstat{\Look{1.00}}{0.00} & \sstat{1.63}{0.06}
                & \sstat{2.95}{0.08} & \sstat{2.48}{0.27}
                & \sstat{1.07}{0.01} & \sstat{1.83}{0.02}
                & \sstat{2.30}{0.03} & \bstat{20.05}{0.36}
        \\
        \textit{ w/o init}
                & \sstat{\Look{1.00}}{0.00} & \sstat{2.36}{0.63}
        		& \sstat{5.18}{0.10} & \sstat{5.72}{1.23}
        		& \sstat{1.45}{0.08} & \sstat{2.29}{0.06}
        		& \sstat{\Look{1.75}}{0.74} & \bstat{29.26}{9.15}
        \\
        \addlinespace
        \textit{Algebraic (\textbf{tree})}
                & \sstat{1.01}{0.00} & \sstat{\Look{1.00}}{0.00}
                & \sstat{\Look{1.05}}{0.01} & \sstat{\Look{1.01}}{0.00}
                & \sstat{\Look{1.00}}{0.00} & \sstat{\Look{1.00}}{0.00}
                & \sstat{2.24}{0.06} & \sstat{\Look{1.83}}{0.02}
        \\
        \textit{ w/o init}
        		& \sstat{1.07}{0.00} & \sstat{\look{1.04}}{0.08}
        		& \sstat{1.44}{0.15} & \sstat{1.27}{0.15}
        		& \sstat{\look{1.05}}{0.10} & \sstat{\Look{1.00}}{0.00}
        		& \sstat{2.42}{0.01} & \sstat{1.86}{0.01}
        \\
		\multicolumn{1}{r@{}}{\smaller\textcolor{gray}{(\textsc{ppl.} / $\downarrow$)}}\\
		\bottomrule
    \end{tabularx}
    \caption{Performance results on algorithmic tree manipulation tasks.}%
    \label{tab:tree_results}
    \end{subtable}\vspace{0.25\baselineskip}
    \begin{subtable}{1\textwidth}
	\smaller
	\centering
	\begin{tabular}{@{~}lcc@{~}}
	& \multicolumn{2}{c}{Epoch}\\
	\cmidrule{2-3}
	Scheme & 
	\multicolumn{1}{c}{$\leq$150}
	&
	\multicolumn{1}{c@{}}{$\leq$300}\\
	\toprule
	~\textit{Sinusoidal 2D} & \sstat{91.57}{0.01} & \sstat{92.79}{0.20}\\
	~\textit{Absolute} & \sstat{90.86}{0.19} & \sstat{92.68}{0.39}\\
	\addlinespace
	~\textit{Algebraic (\textbf{seq})} 
        & \sstat{92.68}{0.24} 
        & \sstat{\look{94.59}}{0.15}
    \\
    ~\textit{ w/o init} 
        & \sstat{88.93}{0.19} 
        & \sstat{91.09}{0.20}
    \\
    \addlinespace
	~\textit{Algebraic (\textbf{grid})} 
        & \sstat{\Look{93.13}}{0.33} 
        & \sstat{\Look{94.67}}{0.06}
    \\
    ~\textit{ w/o init}
        & \sstat{92.95}{0.07} 
        & \sstat{94.48}{0.18}
    \\[-2pt]
	\multicolumn{1}{r@{}}{\smaller\textcolor{gray}{(\textsc{acc.} / $\uparrow$)}}\\
	\bottomrule
	\end{tabular}	
	\caption{Best-by-epoch top-1 accuracy scores on image recognition on \textsc{cifar}-10.}%
	\label{tab:img_results}
    \end{subtable}
    \caption{Experimental results and baselines across the tasks considered.}
    \label{table:results}
\end{table}

\subsection{Tree Transduction}\label{sec:treetrans}
Next, we consider four algorithmic transduction tasks on binary branching trees: tree copying, recursive tree rotation up to a fixpoint, algebraic reduction of C\textsubscript{3} expressions, and self-referential tree manipulation; see Appendix~\ref{appendix:esetup} for details.

In addition to previous sequential baselines, we compare our model to the encodings of \citet[\textit{Tree-SQ}]{shiv2019novel}.
For all four tasks, we experiment with both breadth-first and depth-first decoding.

\subsection{Image Recognition}\label{sec:imagerec}
Finaly, we train a Compact Convolutional Transformer~\citep{hassani2021escaping} on \textsc{cifar}-10~\citep{krizhevsky2009learning}.

Typically, attention-based architectures for vision rely on additive positional encoding schemes, applied on the image prior to it being sequentialized (row-by-row flattened).
Here, we compare fixed~\cite[\textit{Sinusoidal 2D}]{wang2019translating} and parametric~\cite[\textit{Absolute}]{gehring2017convolutional} variants of the above against both the sequential and the grid-structured versions of our scheme.

\subsection{Results}\label{sec:results}
We repeat each experiment three times, varying the seeds used for weight initialization and optimization, but fixing the data across repetitions.
We report means and 95\% CIs in Table~\ref{table:results}.
We highlight each category's best (in red), and underline scores where the CI spans the mean of the respective best.

At the macro level and consistently across modalities, domain-appropriate algebraic interpretations match or surpass strong and specialized baselines -- without \textit{any} hyper-parameter tuning or search.
Specifically, across the 13 setups considered, \ape{} is the uncontested top performer in 8, ranks among the best in 3, and falls within the confidence margin of the top performer in one.
Exceptionally, in the breadth-first version of the tree-copy task, tree algebraic encodings are surpassed by a handful of sequential alternatives; this is no surprise, since in this case the tree structure is practically a task-irrelevant syntactic confound.
Perhaps more surprisingly, in the breadth-first version of the tree-manipulation task, tree algebraic encodings are surpassed only by their non-initialized, sequential version; this is likely a statistical anomaly, since one of the three repetitions resulted in an unusually low perplexity score.

We also note three general trends. 
First, initializing \ape{} to match \rope{} frequency bands at the start of training consistently and significantly improves performance, possibly because \rope{} rotary primitives have undergone empirical tuning for stability and performance. 
Second, given identical initialization, a sequential \ape{} generally outperforms a trainable \rope{}, despite their theoretical equivalence. 
This might be due to the difficulty of optimizing periodic signals (\ie{} \rope{}'s trigonometric functions) compared to \ape{}'s (orthogonal) matrix multiplications. 
Third, a frozen \rope{} performs comparably to a randomly initialized \ape{} in most tasks considered, suggesting that adjusting rotoreflection angles during training is not necessarily better than adjusting rotation planes while keeping the angles fixed.
Contrary to all the above, a frozen \rope{} weakly outperforms both a tunable \rope{} and an initialized \ape{} in the neural machine translation task; likely an artifact of attention overfitting to specific positional patterns.

\section{Related Work}\label{sec:relwork}
Dense attention is by now a foundational component of various problem- and domain-general architectures.
Combined with its structural indifference, this underscores the pressing need for learning strategies capable of injecting structural biases directly at the representation level.
As such, positional encodings have garnered significant community attention in recent years -- too much, in fact, to permit an exhaustive enumeration here.
An extensive survey and meta-review is provided by \citet{dufter-etal-2022-position} who group and rank these works on the basis of several criteria.
Our work presents a universal, intuitive and formally grounded recipe that meets \textit{all} these criteria: it is \textit{trainable}, amenable to problem-specific and data-driven tuning; \textit{reference-adjustable}, allowing both absolute and relative positional specifications; \textit{unbounded}, capable of representing enumerably infinite positions irrespective of model instantiation and/or the targeted data size; \textit{contextual}, implementing a dynamic effect that varies depending on token content;  \textit{effective}, consistently matching or surpassing baselines in the tasks considered; and, finally, \textit{efficient}, exhibiting generally favorable asymptotic complexities.

We must point out that the concept of positional encodings as sequence homomorphisms has already been hinted at, first by \citet{wang2020encoding} and later by \citet{su2023roformer}, even if not explicitly formulated as such.
Despite approaching the problem from different angles, both approaches interpret positions as multiplicative, norm-preserving (rotation-like) operations.
Our proposal expands upon these two, first in providing a proper algebraic framing of the problem, and second in extending the interpretation from rotations around the axes to rotations and reflections about arbitrary planes.
In the case of a single generator matrix (\ie{} sequences), this difference turns to be non-essential, being practically neutralized by the Transformer's trainable weights.
This no longer holds, however, in the case of multiple generator matrices (\ie{} grids or trees), where each generator should be able to rotate and reflect different sets of planes.
In that sense, algebraic positional encodings offer an appealing unifying perspective of a multidimensional generalization to the aforementioned rotation-based frameworks.
This sentiment is shared by \citet{lim2023positional} who, in parallel to our work, similarly advocate for positional encodings as group homomorphisms, there framed as irreducible group representations.
Modulo presentation, the two approaches are variations on a common theme; theirs is technically concerned with post-hoc representation of symmetries and equivariances at a per-datum scale, whereas ours focuses on the interpretation of domain signatures at the dataset scale.

More generally, algebraic manipulations are not uncommon in modern machine learning literature.
The recognition of abstract algebra as a practical tool for imposing structural well-behavedness has led to its increased adoption as a reliable recipe for structure-informed neural architectures, largely obsoleting the inefficient and ad-hoc augmentation routines of the past.
This line of work can be traced back to the group equivariant convolutions of \citet{cohen2016group}, which have by now bloomed into a field of their own; see \citet{weiler2023equivariant} for an up-to-date overview.

\section{Limitations}\label{sec:limitations}
We recognize weaknesses and limitations across three fronts.
On the \textit{theoretical} front, we have limited our scope to simple inductive groups, consciously ignoring potential interpretations of more complex constructions. We defer this to future work.
On the \textit{empirical} front, having to recompute positional encodings once per batch increases a model's temporal complexity during training. While this is barely noticeable in sequential and grid constructions, which scale logarithmically, it becomes evident when dealing with complete trees, which scale linearly and require explicit for-loops.
On the \textit{epistemic} front, we conducted a limited set of experiments, focusing primarily on replicability and fairness. We leave more exhaustive empirical comparisons on practical downstream tasks to future work or interested parties.

\section{Conclusion}
We have presented a theoretically motivated approach towards constructing positional encodings for a variety of structures.
Without any significant modification or overhead, our methodology can capture sequences and their (multi-dimensional as well as multi-branching) generalizations. 
In doing so, it reconciles powerful but structurally oblivious models with their missing inductive biases, permitting structure-aware architectural refinements across a range of tasks and setups (see also \citet{kogkalidis_learning_2024} for parallel work employing the methodology in a neurosymbolic representation learning setup).
Beyond that, our approach grants full control over how these biases are to be implemented, while also being amenable to adjustments and extensions.
Our work indicates that generality and extensibility are not \textit{in spite of}, but rather \textit{due to} structural discipline and abstraction.
We perceive it as an important step towards data-efficient, general and transparent models of neural computation.

\acksection
KK and VG were supported by Saab-WASP via the project ``Neurodynamic Programming and Reinforcement Learning'' (grant 411025). VG also acknowledges the support from Academy of Finland (grant 342077) for ``Human-steered next-generation machine learning for reviving drug design'', and the Jane and Aatos Erkko Foundation (grant 7001703) for ``Biodesign: Use of artificial intelligence in enzyme design for synthetic biology''. 

\bibliography{custom}

\begin{thebibliography}{29}
\providecommand{\natexlab}[1]{#1}
\providecommand{\url}[1]{\texttt{#1}}
\expandafter\ifx\csname urlstyle\endcsname\relax
  \providecommand{\doi}[1]{doi: #1}\else
  \providecommand{\doi}{doi: \begingroup \urlstyle{rm}\Url}\fi

\bibitem[Arjovsky et~al.(2016)Arjovsky, Shah, and Bengio]{arjovsky2016unitary}
M.~Arjovsky, A.~Shah, and Y.~Bengio.
\newblock Unitary evolution recurrent neural networks.
\newblock In \emph{International conference on machine learning}, pages
  1120--1128. PMLR, 2016.

\bibitem[Bernardy and Lappin(2022)]{bernardy2022unitary}
J.-P. Bernardy and S.~Lappin.
\newblock Unitary recurrent networks: Algebraic and linear structures for
  syntax.
\newblock In \emph{Algebraic Structures in Natural Language}, pages 243--278.
  CRC Press, 2022.

\bibitem[Bojar et~al.(2014)Bojar, Buck, Federmann, Haddow, Koehn, Leveling,
  Monz, Pecina, Post, Saint-Amand, et~al.]{bojar2014findings}
O.~Bojar, C.~Buck, C.~Federmann, B.~Haddow, P.~Koehn, J.~Leveling, C.~Monz,
  P.~Pecina, M.~Post, H.~Saint-Amand, et~al.
\newblock Findings of the 2014 workshop on statistical machine translation.
\newblock In \emph{Proceedings of the ninth workshop on statistical machine
  translation}, pages 12--58, 2014.

\bibitem[Cohen and Welling(2016)]{cohen2016group}
T.~Cohen and M.~Welling.
\newblock Group equivariant convolutional networks.
\newblock In \emph{International conference on machine learning}, pages
  2990--2999. PMLR, 2016.

\bibitem[Dufter et~al.(2022)Dufter, Schmitt, and
  Sch{\"u}tze]{dufter-etal-2022-position}
P.~Dufter, M.~Schmitt, and H.~Sch{\"u}tze.
\newblock Position information in transformers: An overview.
\newblock \emph{Computational Linguistics}, 48\penalty0 (3):\penalty0 733--763,
  Sept. 2022.
\newblock \doi{10.1162/coli_a_00445}.
\newblock URL \url{https://aclanthology.org/2022.cl-3.7}.

\bibitem[Gage(1994)]{gage1994new}
P.~Gage.
\newblock A new algorithm for data compression.
\newblock \emph{The C Users Journal}, 12\penalty0 (2):\penalty0 23--38, 1994.

\bibitem[Gehring et~al.(2017)Gehring, Auli, Grangier, Yarats, and
  Dauphin]{gehring2017convolutional}
J.~Gehring, M.~Auli, D.~Grangier, D.~Yarats, and Y.~N. Dauphin.
\newblock Convolutional sequence to sequence learning.
\newblock In \emph{International conference on machine learning}, pages
  1243--1252. PMLR, 2017.

\bibitem[Hassani et~al.(2021)Hassani, Walton, Shah, Abuduweili, Li, and
  Shi]{hassani2021escaping}
A.~Hassani, S.~Walton, N.~Shah, A.~Abuduweili, J.~Li, and H.~Shi.
\newblock Escaping the big data paradigm with compact transformers.
\newblock \emph{arXiv preprint arXiv:2104.05704}, 2021.

\bibitem[Janssen(2014)]{janssen2014foundations}
T.~Janssen.
\newblock \emph{Foundations and applications of Montague grammar}.
\newblock PhD thesis, University of Amsterdam, 2014.
\newblock Originally published: April 1983 (UvA).

\bibitem[Katharopoulos et~al.(2020)Katharopoulos, Vyas, Pappas, and
  Fleuret]{katharopoulos2020transformers}
A.~Katharopoulos, A.~Vyas, N.~Pappas, and F.~Fleuret.
\newblock Transformers are rnns: Fast autoregressive transformers with linear
  attention.
\newblock In \emph{International conference on machine learning}, pages
  5156--5165. PMLR, 2020.

\bibitem[Kogkalidis et~al.(2024)Kogkalidis, Melkonian, and
  Bernardy]{kogkalidis_learning_2024}
K.~Kogkalidis, O.~Melkonian, and J.-P. Bernardy.
\newblock Learning structure-aware representations of dependent types.
\newblock In \emph{The Thirty-eighth Annual Conference on Neural Information
  Processing Systems}, 2024.

\bibitem[Krizhevsky et~al.(2009)Krizhevsky, Hinton,
  et~al.]{krizhevsky2009learning}
A.~Krizhevsky, G.~Hinton, et~al.
\newblock Learning multiple layers of features from tiny images.
\newblock 2009.

\bibitem[Li et~al.(2016)Li, Zemel, Brockschmidt, and Tarlow]{li2016gated}
Y.~Li, R.~Zemel, M.~Brockschmidt, and D.~Tarlow.
\newblock Gated graph sequence neural networks.
\newblock In \emph{Proceedings of ICLR'16}, 2016.

\bibitem[Lim et~al.(2023)Lim, Lawrence, Huang, and Thiede]{lim2023positional}
D.~Lim, H.~Lawrence, N.~T. Huang, and E.~H. Thiede.
\newblock Positional encodings as group representations: A unified framework.
\newblock 2023.

\bibitem[Loshchilov and Hutter(2017)]{loshchilov2017decoupled}
I.~Loshchilov and F.~Hutter.
\newblock Decoupled weight decay regularization.
\newblock \emph{arXiv preprint arXiv:1711.05101}, 2017.

\bibitem[Lu et~al.(2021)Lu, Yao, Zhang, Zhu, Xu, Gao, Xu, Xiang, and
  Zhang]{lu2021soft}
J.~Lu, J.~Yao, J.~Zhang, X.~Zhu, H.~Xu, W.~Gao, C.~Xu, T.~Xiang, and L.~Zhang.
\newblock Soft: Softmax-free transformer with linear complexity.
\newblock \emph{Advances in Neural Information Processing Systems},
  34:\penalty0 21297--21309, 2021.

\bibitem[Murnaghan and Wintner(1931)]{murnaghan1931canonical}
F.~Murnaghan and A.~Wintner.
\newblock A canonical form for real matrices under orthogonal transformations.
\newblock \emph{Proceedings of the National Academy of Sciences}, 17\penalty0
  (7):\penalty0 417--420, 1931.

\bibitem[Post(2018)]{post-2018-call}
M.~Post.
\newblock A call for clarity in reporting {BLEU} scores.
\newblock In \emph{Proceedings of the Third Conference on Machine Translation:
  Research Papers}, pages 186--191, Belgium, Brussels, Oct. 2018. Association
  for Computational Linguistics.
\newblock URL \url{https://www.aclweb.org/anthology/W18-6319}.

\bibitem[Sennrich et~al.(2016)Sennrich, Haddow, and Birch]{sennrich2016neural}
R.~Sennrich, B.~Haddow, and A.~Birch.
\newblock Neural machine translation of rare words with subword units.
\newblock In \emph{Proceedings of the 54th Annual Meeting of the Association
  for Computational Linguistics (Volume 1: Long Papers)}, pages 1715--1725,
  2016.

\bibitem[Shaw et~al.(2018)Shaw, Uszkoreit, and Vaswani]{shaw2018self}
P.~Shaw, J.~Uszkoreit, and A.~Vaswani.
\newblock Self-attention with relative position representations.
\newblock In \emph{Proceedings of NAACL-HLT}, pages 464--468, 2018.

\bibitem[Shiv and Quirk(2019)]{shiv2019novel}
V.~Shiv and C.~Quirk.
\newblock Novel positional encodings to enable tree-based transformers.
\newblock \emph{Advances in neural information processing systems}, 32, 2019.

\bibitem[Su et~al.(2023)Su, Ahmed, Lu, Pan, Bo, and Liu]{su2023roformer}
J.~Su, M.~Ahmed, Y.~Lu, S.~Pan, W.~Bo, and Y.~Liu.
\newblock Roformer: Enhanced transformer with rotary position embedding.
\newblock \emph{Neurocomputing}, page 127063, 2023.

\bibitem[Vaswani et~al.(2017)Vaswani, Shazeer, Parmar, Uszkoreit, Jones, Gomez,
  Kaiser, and Polosukhin]{vaswani2017attention}
A.~Vaswani, N.~Shazeer, N.~Parmar, J.~Uszkoreit, L.~Jones, A.~N. Gomez,
  {\L}.~Kaiser, and I.~Polosukhin.
\newblock Attention is all you need.
\newblock \emph{Advances in neural information processing systems}, 30, 2017.

\bibitem[Vyas et~al.(2020)Vyas, Katharopoulos, and Fleuret]{vyas2020fast}
A.~Vyas, A.~Katharopoulos, and F.~Fleuret.
\newblock Fast transformers with clustered attention.
\newblock \emph{Advances in Neural Information Processing Systems},
  33:\penalty0 21665--21674, 2020.

\bibitem[Wang et~al.(2020)Wang, Donghao, Christina, Li, Peng, Simonsen,
  et~al.]{wang2020encoding}
B.~Wang, Z.~Donghao, L.~Christina, Q.~Li, Z.~Peng, J.~G. Simonsen, et~al.
\newblock Encoding word order in complex embeddings.
\newblock In \emph{ICLR 2020-Proceedings of Eighth International Conference on
  Learning Representations}, 2020.

\bibitem[Wang and Liu(2019)]{wang2019translating}
Z.~Wang and J.-C. Liu.
\newblock Translating math formula images to latex sequences using deep neural
  networks with sequence-level training, 2019.

\bibitem[Weiler et~al.(2023)Weiler, Forr{\'e}, Verlinde, and
  Welling]{weiler2023equivariant}
M.~Weiler, P.~Forr{\'e}, E.~Verlinde, and M.~Welling.
\newblock Equivariant and coordinate independent convolutional networks.
\newblock \emph{A Gauge Field Theory of Neural Networks}, 2023.

\bibitem[Wu et~al.(2021)Wu, Wu, and Huang]{wu2021transformer}
C.~Wu, F.~Wu, and Y.~Huang.
\newblock {DA}-{T}ransformer: Distance-aware transformer.
\newblock In \emph{Proceedings of the 2021 Conference of the North American
  Chapter of the Association for Computational Linguistics: Human Language
  Technologies}, pages 2059--2068, 2021.

\bibitem[Wu et~al.(2016)Wu, Schuster, Chen, Le, Norouzi, Macherey, Krikun, Cao,
  Gao, Macherey, et~al.]{wu2016google}
Y.~Wu, M.~Schuster, Z.~Chen, Q.~V. Le, M.~Norouzi, W.~Macherey, M.~Krikun,
  Y.~Cao, Q.~Gao, K.~Macherey, et~al.
\newblock Google's neural machine translation system: Bridging the gap between
  human and machine translation.
\newblock \emph{arXiv preprint arXiv:1609.08144}, 2016.

\end{thebibliography}
\bibliographystyle{abbrvnat}

\newpage

\appendix
\section{Parameterizing \ape{}}
\label{appendix:init}

\subsection{Orthogonalization}
\label{appendix:orthogonalization}
The orthogonal primitives underlying \ape{} can be procured by matrix-exponentiating skew-symmetric bases. Concretely, for some cyclic group $\langle \mtrx{C} \rangle$:
\begin{enumerate}[topsep=0pt,leftmargin=*,noitemsep]
    \item Start with an \textit{upper triangular} matrix $\mtrx{A}$; this matrix parameterizes the entire group.
    \item Obtain the \textit{skew symmetric} $\mtrx{B} := \mtrx{A} - \mtrx{A}^\top$
    \item Obtain the \textit{matrix exponent} $\mtrx{C} := \mathrm{exp}(\mtrx{B})$; the resulting matrix is \textit{orthogonal}, and acts as the group's generator.
\end{enumerate}

\subsection{Switching between \ape{} and \rope{}}%
\label{appendix:switching}
In the commutative (direct sum of finitely many cyclic groups) case, it is possible to switch freely between \ape{} and \rope{}.
Doing so might be useful, \eg{} for initializing \ape{}, for inspecting the learned rotoreflections post-training, or for making use of \rope{}'s memory-optimized vector-multiplication formula in a system originally trained with \ape{}.
Note that here we consider the purely real-valued version of \rope{} (and \ape{}).

\paragraph{\rope{} $\to$ \ape{}}
To convert \rope{} to \ape{} for some collection of angles $\Theta := [\theta_1, \dots \theta_n]$:
\begin{enumerate}[topsep=0pt,leftmargin=*,noitemsep]
    \item Expand $\Theta$ into a rotation matrix $\mtrx{C}$, 
        \[\mtrx{C} := \begin{bmatrix}
            cos\theta_1 & -sin\theta_1 & 0 & 0 & \dots \\
            sin\theta_1 & cos\theta_1 & 0 & 0 & \dots \\
            0 & 0 & cos\theta_2 & -sin\theta_2 & \dots \\
            0 & 0 & sin\theta_2 & cos\theta_2 & \dots\\
            \vdots & \vdots & \vdots & \vdots & \ddots\\
        \end{bmatrix} \]
    \textbf{Note}: Stop here if not interested in parameterizing $\mtrx{C}$. 
    \item Use a solver to approximate the \textit{matrix logarithm} of $\mtrx{C}$, $\mtrx{B} := \mathrm{log}(\mtrx{C})$.
    \item Find a matrix $\mtrx{A}$ such that $\mathrm{mse}(\mtrx{B}, \mtrx{A}-\mtrx{A}^\top) \leq \epsilon$, \eg{} using a numerical optimizer. Matrix $\mtrx{A}$ can be used to parameterize the group, \textit{cf.} \ref{appendix:orthogonalization}.
\end{enumerate}

\paragraph{\ape{} $\to$ \rope{}} 
To convert \ape{} to \rope{} for some cyclic group $\langle \mtrx{W} \rangle$:
\begin{enumerate}[topsep=0pt,leftmargin=*,noitemsep]
    \item Find the normal form $\mtrx{W} = \mtrx{P}\mtrx{Q}\mtrx{P}^\top$. 
    \item Extract the angles in each block of $\mtrx Q$; the resulting collection of angles is \rope{}'s $\Theta$.
    \item For each attention head involved, right-compose the Transformer's $\mtrx{\Phi}^{(q)}$ and $\mtrx{\Phi}^{(k)}$ with $\mtrx{P}$.
\end{enumerate}

\section{Experimental Setups}
\label{appendix:esetup}

\subsection{Machine Translation}
For our machine translation experiments, we use the official dataset breakdown (including the extended evaluation set).
We tokenize the training and evaluation sets with \textsc{Moses}%
\footnote{See \url{https://github.com/moses-smt/mosesdecoder}}, using the standard pipeline: punctuation normalization $\to$ unicode normalization $\to$ language-specific tokenization.
We apply byte-pair encoding~\citep{gage1994new,sennrich2016neural} using the \texttt{subword-nmt} package%
\footnote{See \url{https://github.com/rsennrich/subword-nmt}.}.
We apply 32k merges across the source and target training corpora, without truncating the resulting (shared) vocabulary (of size 35\,533).
Our loss term is given as the cross-entropy between the teacher-forced predictions and the ground-true labels, smoothed by 10\%.
We train in a distributed environment consisting of 4 GPUs, with a batch size of 3\,072 target tokens per GPU.
We average gradients and update parameters once every 2 GPU iterations (or: 8 batches).
We optimize using Adam with a learning rate dictated by the schedule prescribed by \citet{vaswani2017attention}.
We stop optimizing after 150\,000 parameter updates or 16 hours, whichever comes first.
Throughout training, we circularly store the 10 best checkpoints, ranked on the basis of dev set loss (evaluated once every 500 updates).
During inference, we average the 10 checkpoints into a single model, and select hypotheses from a beam of width 4 and a length penalty of 0.6~\citep{wu2016google}.
We report \textsc{bleu} scores over the \textit{test set} (\texttt{newstest2014}),  comparing the \textsc{bpe}-merged and detokenized output against the raw references using \texttt{sacrebleu}~\citep{post-2018-call}%
\footnote{Signature: \texttt{nrefs:1} | \texttt{case:lc} | \texttt{eff:no} | \texttt{tok:13a} | \texttt{smooth:exp} | \texttt{version:2.4.2}.}.

\begin{table}[h]
	\centering
	\smaller
	\begin{tabular}{@{~}lccc@{~}}
	& \multicolumn{3}{c@{}}{Experiment/Value} \\
    \cmidrule{2-4}
    \multicolumn{1}{@{}l}{Parameter} & NMT & Transduction & Image \\
	\toprule
    Convolution Size        & --            & --               & (3,3)\\
    Convolution Stride      & --            & --               & 1\\
	Embedding Size			& 512           & 512              & 256\\
	Feedforward Size (enc)	& 2048          & 512              & 512\\
	Feedforward Size (dec)	& 2048          & 1024             & --\\
	Feedforward Activation	& ReLU          & ReLU             & GELU\\
	\# Layers (enc, dec)	& (6, 6)        & (2,2)            & (7, 0)\\
	\# Heads				& 8             & 8                & 4\\
	Norm					& LayerNorm     & LayerNorm        & LayerNorm\\
	Norm Position			& Post          & Pre              & Pre\\
    \bottomrule
	\end{tabular}
	\caption{Hyperparameter setups, grouped by experiment.}
	\label{tab:wmt_params}
\end{table}

\subsection{Synthetic Transduction}
\paragraph{Tree Task Descriptions}
The tree copy task is morally identical to its sequential version -- the tree structure (and its positional specification) is practically a confound.

In the tree rotation$^\star$ task, the output tree is the result of recursively right-rotating all subtrees of the input.
The task is challenging but purely structural, in the sense that its resolution requires no real interaction between content and position.

For the algebraic expression reduction task, we consider input trees that specify a complex expression from the cyclic group C\textsubscript{3}, and task the model with producing the result of a single reduction step (\ie{} reducing all subtrees of depth 1 into a leaf).
This time around, the model has to identify reducible subtrees, match operators to their argument and collapse the three into a single node depending on their content.

The tree operations task, finally, combines the aspects of the other three, requiring content-based addressing, structure manipulation and dynamic semantics resolution.
Concretely, we generate an input tree consisting of unique nodes, and randomly select one of its subtrees as well as one of four operators.
We then construct a deeper tree, where the new root corresponds to the chosen operator, its left branch corresponds to the numerical index of the selected subtree, and the right branch is the original tree in its entirety.
The model is then tasked with producing the correct output given this combination of an operator, a tree, and an index.
We consider four operations: extraction (\ie{} return the indexed subtree), flip-extraction (\ie{} return the indexed subtree, rotated), truncation (\ie{} return the full tree with the indexed subtree removed) and a no-op (\ie{} return the full tree as-is, ignoring indexing).

\paragraph{Hyperparameters}
For all synthetic tasks, we generate disjoint train, dev and test sets of sizes 6\,000, 2\,000 and 2\,000.
We train a small Transformer model, optimizing with AdamW~\citep{loshchilov2017decoupled} for 400 epochs and a batch size of 64, using a linear warmup -- cosine decay schedule.
For the sequential tasks, we populate the datasets with words of random lengths from $\mathcal{N}(100, 10)$ and a vocabulary size of $20$ (to ensure token repetition and diffuse the possibility for leaning on content-based addressing).
For the tree tasks, we populate the datasets with non-uniform trees of random depths sampled from $\mathcal{N}(7, 1)$.
For the tree-ops task, exceptionally, we set the vocabulary size to 128 so as to have enough unique nodes to allow content-based addressing.

When using a positional encoding scheme that requires fixing the size of the structure being modeled (\ie{} the \textit{Tree},  \textit{Relative}, and \textit{Absolute} schemes), we fix it at approximately the maximum training size, practically ensuring the most stringent comparison.

In all experiments, we share source and target embedding weights between both the encoder-decoder embedding layers, and the decoder's classification head.

\subsection{Image Recognition}
For our image recognition experiments, we largely rely on the setup of \citet{hassani2021escaping}.
Concretely, we apply a small-step ``tokenizing'' convolution on the input image, downsample the result with max pooling and flatten the result into a sequence. 
After we pass the sequence through the encoder, we apply a global soft attention~\cite[\ia]{li2016gated} (rediscovered by \citet{hassani2021escaping}, there dubbed ``sequence pooling'') to aggregate into a single vector prior to applying the classifier.
To attain competitive scores, we apply standard \textsc{cifar}-10 data augmentations and more aggressive regularization: a 10\% attention weight dropout, a stochastic depth of 10\% for each consecutive layer, and a weight decay of $3\cdot 10^{-2}$.
The above settings and the hyperparameter setup are taken without modification from \citet{hassani2021escaping}.

\newpage
\section*{NeurIPS Paper Checklist}
\begin{enumerate}
\item {\bf Claims}
    \item[] Question: Do the main claims made in the abstract and introduction accurately reflect the paper's contributions and scope?
    \item[] Answer: \answerYes{}
    \item[] Justification: We carefully summarize our contributions and refrain from making any claims that we cannot theoretically or empirically support.
    \item[] Guidelines:
    \begin{itemize}
        \item The answer NA means that the abstract and introduction do not include the claims made in the paper.
        \item The abstract and/or introduction should clearly state the claims made, including the contributions made in the paper and important assumptions and limitations. A No or NA answer to this question will not be perceived well by the reviewers. 
        \item The claims made should match theoretical and experimental results, and reflect how much the results can be expected to generalize to other settings. 
        \item It is fine to include aspirational goals as motivation as long as it is clear that these goals are not attained by the paper. 
    \end{itemize}

\item {\bf Limitations}
    \item[] Question: Does the paper discuss the limitations of the work performed by the authors?
    \item[] Answer: \answerYes{}
    \item[] Justification: We have a dedicated limitations section (\S\ref{sec:limitations}), and openly and explicitly discuss algorithm complexity and experimental scope in the relevant sections.
    \item[] Guidelines:
    \begin{itemize}
        \item The answer NA means that the paper has no limitation while the answer No means that the paper has limitations, but those are not discussed in the paper. 
        \item The authors are encouraged to create a separate "Limitations" section in their paper.
        \item The paper should point out any strong assumptions and how robust the results are to violations of these assumptions (e.g., independence assumptions, noiseless settings, model well-specification, asymptotic approximations only holding locally). The authors should reflect on how these assumptions might be violated in practice and what the implications would be.
        \item The authors should reflect on the scope of the claims made, e.g., if the approach was only tested on a few datasets or with a few runs. In general, empirical results often depend on implicit assumptions, which should be articulated.
        \item The authors should reflect on the factors that influence the performance of the approach. For example, a facial recognition algorithm may perform poorly when image resolution is low or images are taken in low lighting. Or a speech-to-text system might not be used reliably to provide closed captions for online lectures because it fails to handle technical jargon.
        \item The authors should discuss the computational efficiency of the proposed algorithms and how they scale with dataset size.
        \item If applicable, the authors should discuss possible limitations of their approach to address problems of privacy and fairness.
        \item While the authors might fear that complete honesty about limitations might be used by reviewers as grounds for rejection, a worse outcome might be that reviewers discover limitations that aren't acknowledged in the paper. The authors should use their best judgment and recognize that individual actions in favor of transparency play an important role in developing norms that preserve the integrity of the community. Reviewers will be specifically instructed to not penalize honesty concerning limitations.
    \end{itemize}

\item {\bf Theory Assumptions and Proofs}
    \item[] Question: For each theoretical result, does the paper provide the full set of assumptions and a complete (and correct) proof?
    \item[] Answer: \answerYes{}
    \item[] Justification: Our algebraic connections make no assumptions and are fully explicit in their presentation. The equivalence with \rope{} clarifies all assumptions it makes.
    \item[] Guidelines:
    \begin{itemize}
        \item The answer NA means that the paper does not include theoretical results. 
        \item All the theorems, formulas, and proofs in the paper should be numbered and cross-referenced.
        \item All assumptions should be clearly stated or referenced in the statement of any theorems.
        \item The proofs can either appear in the main paper or the supplemental material, but if they appear in the supplemental material, the authors are encouraged to provide a short proof sketch to provide intuition. 
        \item Inversely, any informal proof provided in the core of the paper should be complemented by formal proofs provided in appendix or supplemental material.
        \item Theorems and Lemmas that the proof relies upon should be properly referenced. 
    \end{itemize}

    \item {\bf Experimental Result Reproducibility}
    \item[] Question: Does the paper fully disclose all the information needed to reproduce the main experimental results of the paper to the extent that it affects the main claims and/or conclusions of the paper (regardless of whether the code and data are provided or not)?
    \item[] Answer: \answerYes{}
    \item[] Justification: We provide the reviewers with both an extensive appendix detailing our experimental setup, and the code used to implement our methodology and conduct our experiments.
    \item[] Guidelines:
    \begin{itemize}
        \item The answer NA means that the paper does not include experiments.
        \item If the paper includes experiments, a No answer to this question will not be perceived well by the reviewers: Making the paper reproducible is important, regardless of whether the code and data are provided or not.
        \item If the contribution is a dataset and/or model, the authors should describe the steps taken to make their results reproducible or verifiable. 
        \item Depending on the contribution, reproducibility can be accomplished in various ways. For example, if the contribution is a novel architecture, describing the architecture fully might suffice, or if the contribution is a specific model and empirical evaluation, it may be necessary to either make it possible for others to replicate the model with the same dataset, or provide access to the model. In general. releasing code and data is often one good way to accomplish this, but reproducibility can also be provided via detailed instructions for how to replicate the results, access to a hosted model (e.g., in the case of a large language model), releasing of a model checkpoint, or other means that are appropriate to the research performed.
        \item While NeurIPS does not require releasing code, the conference does require all submissions to provide some reasonable avenue for reproducibility, which may depend on the nature of the contribution. For example
        \begin{enumerate}
            \item If the contribution is primarily a new algorithm, the paper should make it clear how to reproduce that algorithm.
            \item If the contribution is primarily a new model architecture, the paper should describe the architecture clearly and fully.
            \item If the contribution is a new model (e.g., a large language model), then there should either be a way to access this model for reproducing the results or a way to reproduce the model (e.g., with an open-source dataset or instructions for how to construct the dataset).
            \item We recognize that reproducibility may be tricky in some cases, in which case authors are welcome to describe the particular way they provide for reproducibility. In the case of closed-source models, it may be that access to the model is limited in some way (e.g., to registered users), but it should be possible for other researchers to have some path to reproducing or verifying the results.
        \end{enumerate}
    \end{itemize}

\item {\bf Open access to data and code}
    \item[] Question: Does the paper provide open access to the data and code, with sufficient instructions to faithfully reproduce the main experimental results, as described in supplemental material?
    \item[] Answer: \answerYes{}
    \item[] Justification: Yes -- see answer above. Our training scripts are provided virtually unchanged.
    \item[] Guidelines:
    \begin{itemize}
        \item The answer NA means that paper does not include experiments requiring code.
        \item Please see the NeurIPS code and data submission guidelines (\url{https://nips.cc/public/guides/CodeSubmissionPolicy}) for more details.
        \item While we encourage the release of code and data, we understand that this might not be possible, so “No” is an acceptable answer. Papers cannot be rejected simply for not including code, unless this is central to the contribution (e.g., for a new open-source benchmark).
        \item The instructions should contain the exact command and environment needed to run to reproduce the results. See the NeurIPS code and data submission guidelines (\url{https://nips.cc/public/guides/CodeSubmissionPolicy}) for more details.
        \item The authors should provide instructions on data access and preparation, including how to access the raw data, preprocessed data, intermediate data, and generated data, etc.
        \item The authors should provide scripts to reproduce all experimental results for the new proposed method and baselines. If only a subset of experiments are reproducible, they should state which ones are omitted from the script and why.
        \item At submission time, to preserve anonymity, the authors should release anonymized versions (if applicable).
        \item Providing as much information as possible in supplemental material (appended to the paper) is recommended, but including URLs to data and code is permitted.
    \end{itemize}

\item {\bf Experimental Setting/Details}
    \item[] Question: Does the paper specify all the training and test details (e.g., data splits, hyperparameters, how they were chosen, type of optimizer, etc.) necessary to understand the results?
    \item[] Answer: \answerYes{}
    \item[] Justification: Yes, see above.
    \item[] Guidelines:
    \begin{itemize}
        \item The answer NA means that the paper does not include experiments.
        \item The experimental setting should be presented in the core of the paper to a level of detail that is necessary to appreciate the results and make sense of them.
        \item The full details can be provided either with the code, in appendix, or as supplemental material.
    \end{itemize}

\item {\bf Experiment Statistical Significance}
    \item[] Question: Does the paper report error bars suitably and correctly defined or other appropriate information about the statistical significance of the experiments?
    \item[] Answer: \answerYes{}
    \item[] Justification: We take extra care to conduct our experiments openly and transparently so as to deliver statistically sound results and draw solid conclusions. We repeat \textit{all} experiments multiple times, and report means and 95\% confidence intervals. For each experiment, we visually mark all models that overlap with the best performer in the category.
    \item[] Guidelines:
    \begin{itemize}
        \item The answer NA means that the paper does not include experiments.
        \item The authors should answer "Yes" if the results are accompanied by error bars, confidence intervals, or statistical significance tests, at least for the experiments that support the main claims of the paper.
        \item The factors of variability that the error bars are capturing should be clearly stated (for example, train/test split, initialization, random drawing of some parameter, or overall run with given experimental conditions).
        \item The method for calculating the error bars should be explained (closed form formula, call to a library function, bootstrap, etc.)
        \item The assumptions made should be given (e.g., Normally distributed errors).
        \item It should be clear whether the error bar is the standard deviation or the standard error of the mean.
        \item It is OK to report 1-sigma error bars, but one should state it. The authors should preferably report a 2-sigma error bar than state that they have a 96\% CI, if the hypothesis of Normality of errors is not verified.
        \item For asymmetric distributions, the authors should be careful not to show in tables or figures symmetric error bars that would yield results that are out of range (e.g. negative error rates).
        \item If error bars are reported in tables or plots, The authors should explain in the text how they were calculated and reference the corresponding figures or tables in the text.
    \end{itemize}

\item {\bf Experiments Compute Resources}
    \item[] Question: For each experiment, does the paper provide sufficient information on the computer resources (type of compute workers, memory, time of execution) needed to reproduce the experiments?
    \item[] Answer: \answerNo{} 
    \item[] Justification: While we do report hardware infrastructure, we do not report memory consumption or clock times. With the exception of machine translation, our experiments are moderately cheap to run, requiring no specialized hardware other than GPU accelaration.
    \item[] Guidelines:
    \begin{itemize}
        \item The answer NA means that the paper does not include experiments.
        \item The paper should indicate the type of compute workers CPU or GPU, internal cluster, or cloud provider, including relevant memory and storage.
        \item The paper should provide the amount of compute required for each of the individual experimental runs as well as estimate the total compute. 
        \item The paper should disclose whether the full research project required more compute than the experiments reported in the paper (e.g., preliminary or failed experiments that didn't make it into the paper). 
    \end{itemize}
    
\item {\bf Code Of Ethics}
    \item[] Question: Does the research conducted in the paper conform, in every respect, with the NeurIPS Code of Ethics \url{https://neurips.cc/public/EthicsGuidelines}?
    \item[] Answer: \answerYes{}
    \item[] Justification: Checked and done.
    \item[] Guidelines:
    \begin{itemize}
        \item The answer NA means that the authors have not reviewed the NeurIPS Code of Ethics.
        \item If the authors answer No, they should explain the special circumstances that require a deviation from the Code of Ethics.
        \item The authors should make sure to preserve anonymity (e.g., if there is a special consideration due to laws or regulations in their jurisdiction).
    \end{itemize}

\item {\bf Broader Impacts}
    \item[] Question: Does the paper discuss both potential positive societal impacts and negative societal impacts of the work performed?
    \item[] Answer: \answerYes{}
    \item[] Justification: We do, albeit briefly. We do not see possible negative implications.
    \item[] Guidelines:
    \begin{itemize}
        \item The answer NA means that there is no societal impact of the work performed.
        \item If the authors answer NA or No, they should explain why their work has no societal impact or why the paper does not address societal impact.
        \item Examples of negative societal impacts include potential malicious or unintended uses (e.g., disinformation, generating fake profiles, surveillance), fairness considerations (e.g., deployment of technologies that could make decisions that unfairly impact specific groups), privacy considerations, and security considerations.
        \item The conference expects that many papers will be foundational research and not tied to particular applications, let alone deployments. However, if there is a direct path to any negative applications, the authors should point it out. For example, it is legitimate to point out that an improvement in the quality of generative models could be used to generate deepfakes for disinformation. On the other hand, it is not needed to point out that a generic algorithm for optimizing neural networks could enable people to train models that generate Deepfakes faster.
        \item The authors should consider possible harms that could arise when the technology is being used as intended and functioning correctly, harms that could arise when the technology is being used as intended but gives incorrect results, and harms following from (intentional or unintentional) misuse of the technology.
        \item If there are negative societal impacts, the authors could also discuss possible mitigation strategies (e.g., gated release of models, providing defenses in addition to attacks, mechanisms for monitoring misuse, mechanisms to monitor how a system learns from feedback over time, improving the efficiency and accessibility of ML).
    \end{itemize}
    
\item {\bf Safeguards}
    \item[] Question: Does the paper describe safeguards that have been put in place for responsible release of data or models that have a high risk for misuse (e.g., pretrained language models, image generators, or scraped datasets)?
    \item[] Answer: \answerNA{}
    \item[] Justification: We perceive no risks that would require safeguards of any kind.
    \item[] Guidelines:
    \begin{itemize}
        \item The answer NA means that the paper poses no such risks.
        \item Released models that have a high risk for misuse or dual-use should be released with necessary safeguards to allow for controlled use of the model, for example by requiring that users adhere to usage guidelines or restrictions to access the model or implementing safety filters. 
        \item Datasets that have been scraped from the Internet could pose safety risks. The authors should describe how they avoided releasing unsafe images.
        \item We recognize that providing effective safeguards is challenging, and many papers do not require this, but we encourage authors to take this into account and make a best faith effort.
    \end{itemize}

\item {\bf Licenses for existing assets}
    \item[] Question: Are the creators or original owners of assets (e.g., code, data, models), used in the paper, properly credited and are the license and terms of use explicitly mentioned and properly respected?
    \item[] Answer: \answerYes{}
    \item[] Justification: We cite all software libraries and datasets we use, and comply with their licenses.
    \item[] Guidelines:
    \begin{itemize}
        \item The answer NA means that the paper does not use existing assets.
        \item The authors should cite the original paper that produced the code package or dataset.
        \item The authors should state which version of the asset is used and, if possible, include a URL.
        \item The name of the license (e.g., CC-BY 4.0) should be included for each asset.
        \item For scraped data from a particular source (e.g., website), the copyright and terms of service of that source should be provided.
        \item If assets are released, the license, copyright information, and terms of use in the package should be provided. For popular datasets, \url{paperswithcode.com/datasets} has curated licenses for some datasets. Their licensing guide can help determine the license of a dataset.
        \item For existing datasets that are re-packaged, both the original license and the license of the derived asset (if it has changed) should be provided.
        \item If this information is not available online, the authors are encouraged to reach out to the asset's creators.
    \end{itemize}

\item {\bf New Assets}
    \item[] Question: Are new assets introduced in the paper well documented and is the documentation provided alongside the assets?
    \item[] Answer: \answerNo{}
    \item[] Justification: While we do provide reference implementations, we do not see them as assets per se, neither do we hand them out as ready-to-use integrations.
    \item[] Guidelines:
    \begin{itemize}
        \item The answer NA means that the paper does not release new assets.
        \item Researchers should communicate the details of the dataset/code/model as part of their submissions via structured templates. This includes details about training, license, limitations, etc. 
        \item The paper should discuss whether and how consent was obtained from people whose asset is used.
        \item At submission time, remember to anonymize your assets (if applicable). You can either create an anonymized URL or include an anonymized zip file.
    \end{itemize}

\item {\bf Crowdsourcing and Research with Human Subjects}
    \item[] Question: For crowdsourcing experiments and research with human subjects, does the paper include the full text of instructions given to participants and screenshots, if applicable, as well as details about compensation (if any)? 
    \item[] Answer: \answerNA{} 
    \item[] Justification: No human subjects were involved in this study.
    \item[] Guidelines:
    \begin{itemize}
        \item The answer NA means that the paper does not involve crowdsourcing nor research with human subjects.
        \item Including this information in the supplemental material is fine, but if the main contribution of the paper involves human subjects, then as much detail as possible should be included in the main paper. 
        \item According to the NeurIPS Code of Ethics, workers involved in data collection, curation, or other labor should be paid at least the minimum wage in the country of the data collector. 
    \end{itemize}

\item {\bf Institutional Review Board (IRB) Approvals or Equivalent for Research with Human Subjects}
    \item[] Question: Does the paper describe potential risks incurred by study participants, whether such risks were disclosed to the subjects, and whether Institutional Review Board (IRB) approvals (or an equivalent approval/review based on the requirements of your country or institution) were obtained?
    \item[] Answer: \answerNA{}
    \item[] Justification: See above.
    \item[] Guidelines:
    \begin{itemize}
        \item The answer NA means that the paper does not involve crowdsourcing nor research with human subjects.
        \item Depending on the country in which research is conducted, IRB approval (or equivalent) may be required for any human subjects research. If you obtained IRB approval, you should clearly state this in the paper. 
        \item We recognize that the procedures for this may vary significantly between institutions and locations, and we expect authors to adhere to the NeurIPS Code of Ethics and the guidelines for their institution. 
        \item For initial submissions, do not include any information that would break anonymity (if applicable), such as the institution conducting the review.
    \end{itemize}

\end{enumerate}

\end{document}